\documentclass[10pt,twocolumn,letterpaper]{article}

\usepackage{iccv}
\usepackage{times}
\usepackage{epsfig}
\usepackage{graphicx}
\usepackage{amsmath}
\usepackage{amssymb}

\usepackage[pagebackref=true,breaklinks=true,letterpaper=true,colorlinks,bookmarks=false]{hyperref}
\usepackage{algorithmic}
\usepackage{algorithm}
\usepackage{multirow}
\usepackage{comment}
\usepackage{threeparttable}

\iccvfinalcopy 

\def\iccvPaperID{4120} 

\ificcvfinal\fi

\begin{document}

\title{$n$-hot: Efficient Bit-Level Sparsity for Powers-of-Two\\ 
Neural Network Quantization}
\author{
   Yuiko Sakuma, Hiroshi Sumihiro, Jun Nishikawa, Toshiki Nakamura, Ryoji Ikegaya\\
   {\tt\small \{Yuiko.Sakuma, Hiroshi.Sumihiro, Jun.B.Nishikawa, Toshiki.A.Nakamura, Ryoji.Ikegaya\}@sony.com}\\
   Sony Corporate, Japan\\
   1-7-1 Konan Minato-ku, Tokyo, 108-0075 Japan\\
}

\maketitle
\ificcvfinal\thispagestyle{empty}\fi

\begin{abstract}
   Powers-of-two (PoT) quantization reduces the number of bit operations of deep neural networks on resource-constrained hardware.
   However, PoT quantization triggers a severe accuracy drop because of its limited representation ability. Since DNN models have been applied for relatively complex tasks (e.g., classification for large datasets and object detection), improvement in accuracy for the PoT quantization method is required.
   Although some previous works attempt to improve the accuracy of PoT quantization, there is no work that balances accuracy and computation costs in a memory-efficient way.
   To address this problem, we propose an efficient PoT quantization scheme. Bit-level sparsity is introduced; weights (or activations) are rounded to values that can be calculated by $n$ shift operations in multiplication. 
   We also allow not only addition but also subtraction as each operation. 
   Moreover, we use a two-stage fine-tuning algorithm to recover the accuracy drop that is triggered by introducing the bit-level sparsity. 
   The experimental results on an object detection model (CenterNet, MobileNet-v2 backbone) on the COCO dataset show that our proposed method suppresses the accuracy drop by 0.3\% at most while reducing the number of operations by about 75\% and model size by 11.5\% compared to the uniform method.
\end{abstract}

\section{Introduction}

Deep neural networks (DNNs) have shown a significant improvement in performance for various tasks and real-life applications. 
However, recently proposed DNNs tend to be over-parameterized and difficult to implement on resource-constrained devices (e.g., smartwatches, etc.) \cite{Han2015, Wen2016}.
To address this challenge, various DNN compression techniques including quantization \cite{Zhou2016, Zhou2017}, pruning \cite{Han2015a,Molchanov2019}, and other compression techniques including low-rank decomposition \cite{Denil2013}, knowledge distillation (KD) \cite{Hinton2015}, and neural architecture search \cite{Wang2019haq} have been proposed. 

For efficient hardware (HW) deployment, fixed-point quantization \cite{Lin2015} has been widely studied.
Fixed-point quantization is often applied to real HW deployment instead of floating-point bit expression because it can reduce memory usage.
When DNN is operated on HW, the output of the network is calculated by multiplying activations and weights.
However, the computation cost for multiplication is large on HW \cite{Chen2020};
reducing computation cost for multiplication operation efficiently speeds up DNN operation on HW.
Coubariaux et al. \cite{Courbariaux2015} proposed BinaryConnect that binarizes weights. 
On binary networks, the calculation can be accelerated because multiplication operations can be replaced by simple accumulation.
Coubariaux et al. \cite{Courbariaux2016} and Hubura et al. \cite{Hubara2016} proposed BinaryNet that expanded the previous studies; not only weights but activations are also binarized.
While BinaryNet suffered from severe accuracy loss for larger datasets, Rastegari et al. \cite{Rastegari2016} adjusted binarized values per layer to improve recognition accuracy for ImageNet dataset \cite{Deng2009}. 

However, since binary representation has a lower capacity of information, the accuracy tends to be low. 
Another approach to reducing the computation cost on multiplication is considering the shift and additive operation method.
Multiplication operation can be decomposed into shift and addition operation and actual DNN calculation on HW can be achieved by these. 
Chen et al. \cite{Chen2020} proposed AdderNet that only required additive operations. 
The convolution layers are replaced with proposed AdderNet layers that only use additive operations.
Although AdderNet benefits from limited operations, it may suffer from accuracy drop for certain tasks because it cannot use conventional convolution layers. 

On the other hand, Miyashita et al. \cite{Miyashita2016} expressed weights by only using values with powers-of-two (PoT) quantization levels.
PoT quantization reduces the number of shifts in multiplication.
For example, when operating a typical DNN with 8-bit weights, eight times shift operations are required.
However, if the weights are expressed by PoT values, the shift operation occurs once.
While PoT quantization can largely reduce computation costs, the expressible values are limited.
For instance, while quantization levels near zero are dense, they are sparse for values near the maximum value because the step size increases by $\times 2$.
This leads to the degradation of the quantized DNNs' recognition ability.
Especially, accuracy loss is severe for quantization for complex tasks such as object detection. 

Li et al. \cite{Li2020} expands PoT representation by allowing $n > 1$ shift operations in multiplication. 
They proposed additive PoT (APoT) \cite{Li2020} that expresses values by adding PoT terms. 
The experimental results show that $n = 2$ APoT quantization largely improves classification accuracy while the number of shift operations is limited to two times; compared to a PoT representation, APoT can express values near the maximum value of the quantization range.
Although APoT is a powerful method to balance computation cost and accuracy, it does not consider model size reduction. 
Reducing memory is important for memory-restricted devices.
Therefore, a memory-efficient quantization method that balances reduction in the number of shift operations and accuracy is required.

We propose an ``$n$-hot'' quantization for DNNs.
Different from previous approaches, we introduce bit-level sparsity for efficient weight representation to address the tradeoff. 
Our main contributions are as follows:
\begin{itemize}
   \item  We introduce {\it bit-level sparsity}; 
   each weight (or activation) expressed by less than or equal to  $n$ PoT terms by selecting the appropriate values from the original $b$-bit precision model, thus reducing the model size.
   The proposed $n$-hot quantization reduces the number of operations in multiplication that for DNNs with $b$-bit weights, that $n/b$ bit operations are required than that of uniform quantization schemes.
   Moreover, we use not only addition but {\it subtraction} of PoT terms (we call this bit representation ``$n$-hot'').
   By considering subtraction, the accuracy improves without increasing the number of operations because the number of expressible values increases.
   \item For fine-tuning, we show a {\it two-stage fine-tuning algorithm} - firstly fine-tunes by only quantizing activations, then secondly fine-tunes by quantizing both activations and weights - effectively recovers the accuracy drop that is triggered by introducing bit-level sparsity for object detection tasks.
   \item Our proposed method is evaluated on different tasks (e.g., classification and object detection) and settings (e.g., quantization on pruned networks) on MobileNet and ResNet models by using the COCO and CIFAR100 datasets.
   Specifically, compared to the uniform method, our proposed method suppresses the accuracy drop by 0.4\% while reducing the number of bit operations by approximately 75\% and the model size by 11.5\% for 8-bit quantization on the COCO dataset.
\end{itemize}

\section{Related Works}

\subsection{Non-uniform quantization}
Several works study non-uniform quantization schemes. 
Seo et al. \cite{Seo2019} used kernel density estimation based non-uniform quantizer. 
While some works use deterministic methods, others use learnable functions to optimize quantizers.
Zhang et al. \cite{Zhang2018} proposed LQ-Nets, which learns quantizers for both weights and activations. 
Polino et al. \cite{Polino2018} introduced a differentiable quantization method that optimizes the location of quantization points by using a KD technique.
Although using learnable non-uniform quantization levels improves accuracy, these works do not consider computation costs on HW.

PoT quantization schemes \cite{Miyashita2016, Zhou2017} use logarithmic quantization methods. 
Because of its efficiency on HW, several works \cite{Ueyoshi2019, Xu2020} applied PoT to accelerate multiplication operation on HW for DNNs.
Several works seek to improve the performance of DNNs that are quantized with PoT values.
Vogel et al. \cite{Vogel2018} optimized base values of the quantizer to improve accuracy. 
As explained in the previous section, Li et al. \cite{Li2020} proposed APoT that addresses the tradeoff between accuracy and the number of operations.
However, no previous works propose a memory-efficient PoT scheme to balance accuracy and computation costs.

\subsection{Quantization on complex tasks}
Recent studies on DNN quantization mostly fall into two categories.
One achieves extremely low-bit such as ternary \cite{Li2016, Zhou2016} or even binary \cite{Courbariaux2015, Hubara2016,Rastegari2016} compaction.
These extremely low-bit quantization techniques are mostly performed on relatively simple tasks such as classification.
Another category is quantization techniques for relatively complex tasks such as natural language processing \cite{Bhandare2019}, semantic segmentation \cite{Zhuang2019}, and object detection \cite{Yin2017, Li2019, Jacob2018}.

Especially, object detection has been intensively studied because of its wide range of applications (e.g., surveillance \cite{Xu2018} and autonomous driving \cite{Wang2019a}).
There are various object detection techniques including two stage (\cite{Ren2015}) and one-stage (\cite{Liu2016,Redmon2016,Lin2015a}) methods.
Among previously proposed methods, CenterNet \cite{Zhou2019} is a one-stage, simple, and efficient method; it represents objects as a single point at their bounding box center.
It achieved the best speed-accuracy trade-off on the MS COCO dataset \cite{COCO}.

Several works study quantization techniques on object detection tasks.
Jacob et al. \cite{Jacob2018} evaluated their quantization scheme on SSD \cite{Liu2016} and showed a small accuracy drop when quantized to 8-bit.
Yin et al. \cite{Yin2017} proposed LBW-Net, which efficiently optimizes quantized network by using full-precision weights.
Although most previous works study larger bit precision, such as 8-bit, Li et al. \cite{Li2019} further quantized RetinaNet \cite{Lin2015a} and Faster R-CNN \cite{Ren2015} to 4-bit.
However, no previous work studies quantization on CenterNet.
Moreover, since object detection has been an important task for real-life application, a study on the performance of non-uniformly quantized DNN models for object detection is required for their HW implementation.

\section{Proposed Method}

\begin{figure}[t]
\begin{center}
\fbox{
   \includegraphics[bb=0 0 260 130, width=0.9\linewidth]
   {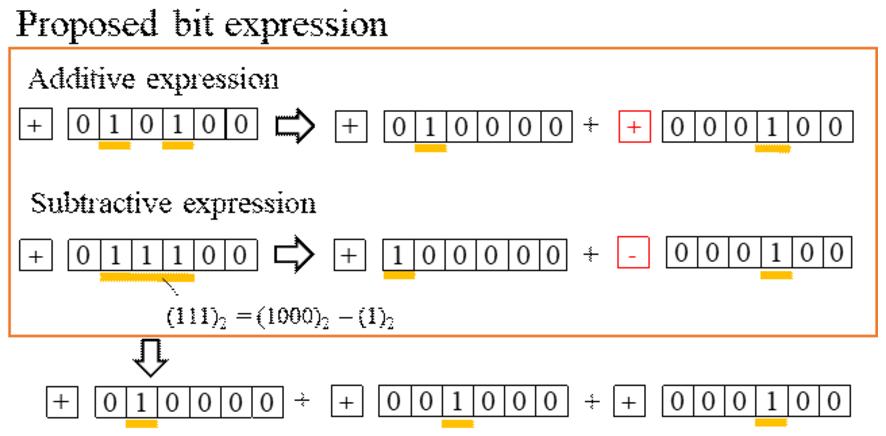}
   }
   \caption{Concept of the proposed $n$-hot quantization scheme.}
\label{fig:concept}
\end{center}
\end{figure}

\subsection{$n$-hot quantization}

Our proposed quantization scheme aims to select appropriate values that can be calculated by the limited number of shift operations during multiplying weights and activations without a large accuracy loss.
We take the approach of selecting values that can be calculated by less than or equal to $n$ shifts.
For instance, weights (or activations) are rounded to the values that can be expressed by $n$ PoT terms.

Considering a binary expression, values are expressed as the addition of different PoT values (e.g., one-hot values).
When two PoT terms are added (e.g., $(010000)_2 + (000100)_2 = (010100)_2$), the required number of the shift operations in multiplication is two.
Our proposed bit expression further considers the case of {\it subtraction} to expand the representation ability. 
More than two consecutive ``1''s in the bit expression can be replaced by subtractive expression.
For example, ``$(011100)_2$'' can be replaced by ``$(100000)_2 - (000100)_2$''.
Introducing the subtractive expression, the number of the expressible values increases which leads to improvement in the accuracy of the quantized DNN. 
For example, for 8-bit, when using only addition, the number of expressible values is 37; considering subtraction, it is expanded to 58.

Figure \ref{fig:concept} briefly explains the bit expression of our $n$-hot quantization scheme for the case of $n = 2$ and bit-width $b = 6$.
The values are expressed by either addition or subtraction of different PoT values of $b$-bit.
Subtractive expression effectively increases the number of expressible values.
For instance, if the bit expression $(011100)_2$ is expressed by addition, it is decomposed to $(010000)_2 + (001000)_2 + (000100)_2$, which required three shift operations.
However, if a subtractive unit is implemented on HW, it can be operated by only two shift operations.

\begin{figure*}
\begin{center}
\fbox{
   \includegraphics[bb= 0 0 430 160, width=0.9\linewidth]{
      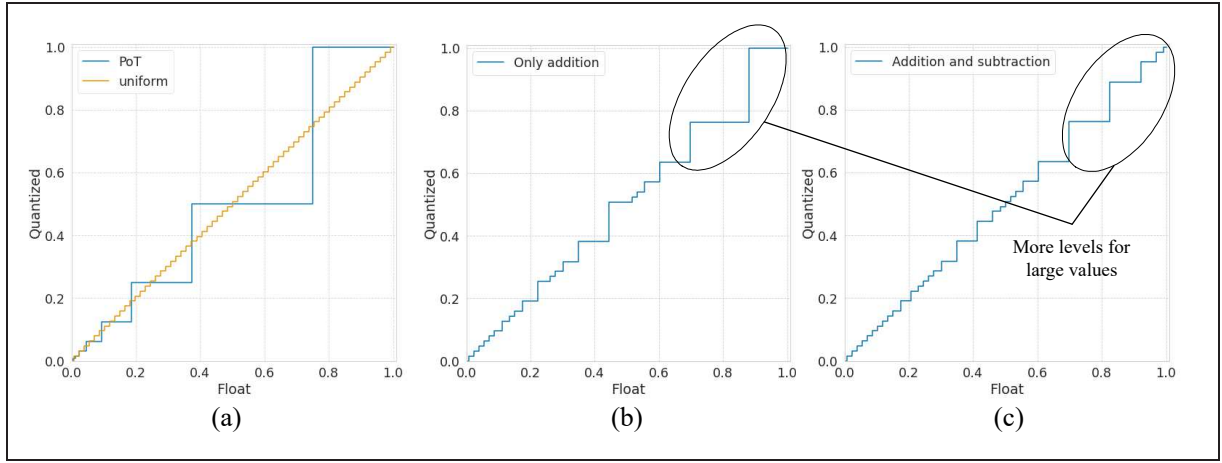}
}
   \caption{Quantized values for 
   (a) PoT and uniform (b) only addition and (c) addition and subtraction (proposed).}
\label{fig:bit_expressions}
\end{center}
\end{figure*}

Figure \ref{fig:bit_expressions} depicts the difference in the number of expressible values for different quantization methods when $b=5$.
For simplicity, only the absolute values are presented, and values are quantized between 0 and 1 for this example; 
although the discussions are made for positive values, negative values have the same characteristics.
Near zero, both ``Only addition'' (Figure \ref{fig:bit_expressions}(b)) and ``Addition and subtraction'' (Figure \ref{fig:bit_expressions}(c)) methods express the same quantized levels as the uniform case. 
Since weight distributions of typical DNNs are bell-shaped, most weight values are near zero, this prevents ``Only addition'' and ``Addition and subtraction'' (e.g., our proposed $n$-hot quantization) methods from degrading their accuracies; 
most values can be expressed by the same quantization levels as the original bit expression.
Near the maximum value, while ``Only addition'' has large quantization levels, ``Addition and subtraction'' has smaller quantization levels.
Comparing to ``Only addition'', ``Addition and subtraction'' can improve the quantized resolution around the maximum value.
This fact suppresses accuracy drop especially for complex tasks that require smaller quantization levels (e.g., DNN models for object detection) without increasing computation costs.

Figure \ref{fig:bit_expressions}(a) also illustrates bit representation for the conventional PoT quantization (e.g., values between 0 and 1 are expressed by quantization levels 0, $2^{-2^b-1}$,…,  $2^{-1}$, 1).
Comparing to the proposed representation (Figure \ref{fig:bit_expressions}(c)), while the quantization level around zero is small, it is much broader for values larger than around 0.4.
It causes a large approximation loss for the quantized DNN, and which results in an accuracy drop.
Comparing to the conventional PoT, our proposed method increases the number of quantization levels while guaranteeing the number of shift operations to $n$. 

The selective values for our proposed $n$-hot quantization can be generalized in Equation (\ref{eq:generalized_nhot}) for $b$-bit quantized weights when values are quantized between 0 and 1:
\begin{align}
   &P_b = \{2^0, 2^{-1}, 2^{-2}, ..., 2^{-(b-1)}\} \label{eq:PoT_values} \\
   &Q_{n-hot}(\alpha, b, n) = \nonumber\\
    &\{
    \{\alpha \sum_{i=1}^{n} (-1)^{c_i}P_{b,i}\} \cup
    \{\alpha \sum_{i=1}^{n-1} (-1)^{c_i}P_{b,i}\} \cup
    ... \nonumber \\
    &\cup \{\alpha \sum_{i=1}^{2} (-1)^{c_i}P_{b,i}\} \cup
    \{\alpha (-1)^{c_i}P_{b, i}\} \cup
    \{0\}| \nonumber \\ 
    &P_{b,1} < P_{b,2},... < P_{b,n}, c_i \in \{0, 1\},
    \alpha > 0
    \} 
   \label{eq:generalized_nhot}
\end{align}
where $\alpha$ is a scale.
Equation (\ref{eq:PoT_values}) represents PoT values for $b$ bit.
For any real value $x_{input}$, $x_{input}$ is projected by $\Pi(\cdot)$ to $n$-hot quantization levels $Q_{n-hot}(\alpha, b, n)$ (Equation \ref{eq:nhot_projection}).
In this paper, $\Pi(\cdot)$ is the projection to the nearest value. 
\begin{equation}
   x_{quantized} = \Pi_{Q_{n-hot}(\alpha, b, n)(x_{input})}
   \label{eq:nhot_projection}
\end{equation}

For example, in the case of $n = 2$, Equation \ref{eq:generalized_nhot} can be expanded as follows:
\begin{align}
   Q_{n-hot}&(\alpha, b, n = 2) = \nonumber \\
   &\{\alpha (P_{b,1} + P_{b,2})\} \cup
   \{\alpha (P_{b,1} - P_{b,2})\} \cup \nonumber \\
   &\{\alpha (-P_{b,1} - P_{b,2})\} \cup
   \{\alpha (-P_{b,1} + P_{b,2})\} \cup \nonumber \\
   & \{\alpha P_{b}\} \cup \{\alpha (-P_{b})\} \cup \{0\}
   \label{eq:2-hot_example}
\end{align}
Similar expansion can be performed for other $n$.


\subsection{Computation on HW}

\begin{figure*}
\begin{center}
\fbox{
   \includegraphics[bb= 0 0 830 360, width=0.9\linewidth]{
      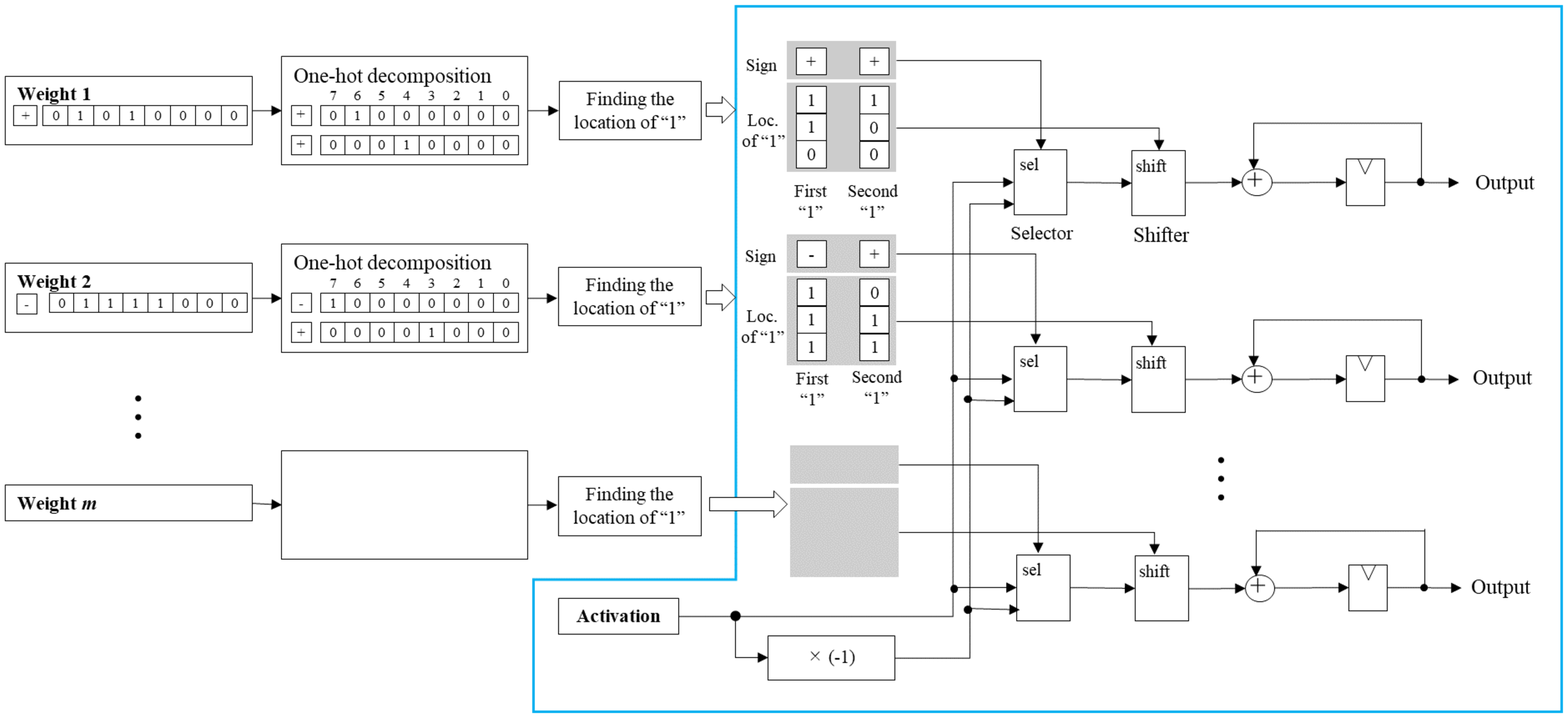}
}
   \caption{HW architecture for the proposed method.}
\label{fig:circuit}
\end{center}
\end{figure*}

The possible HW architecture is illustrated in Figure \ref{fig:circuit}.
Figure \ref{fig:circuit} presents the case that each weight is expressed by the proposed $n$-hot quantization with 1 bit for a sign and 8-bit for an absolute value, and each activation is expressed by a conventional uniform quantization representation.
The input weights are rounded to values that can be expressed by the proposed $n$-hot quantization. Multiplication is performed by $n$ times shifts and additions to get the output activation.

Each weight is decomposed to one-hot values with signs.
Then, the location of ``1'' is searched and kept in some memory units such as a flip-flop.
Selectors retrieve positive or negative activation values depending on the sign of the weight.
Shift operations are performed depending on the information of the location of ``1''.
Finally, the shifted values are accumulated by $n$ times to get the output.

Although in Figure \ref{fig:circuit}, additional memories to keep the location of ``1'' (indicated by grey color) and circuits to decompose inputs and finding the locations of ``1'' are indicated, these can be omitted; in an offline process, the sign and location information can be retrieved in advance. In this case, the circuit which reads out the sign and location from the memory (indicated by the blue box) only is required.

The HW architecture can also be applied for the opposite case when activations and weights are represented by the proposed and the conventional uniform schemes, respectively.
In that case, simply, the locations of weights and activations are switched in Figure \ref{fig:circuit}.


\subsection{Relation to other PoT quantization schemes}

When $n = 1$, our proposed bit expression can be described as a special case of PoT, that is, the bit expression is PoT, but the number of possible bit values is only $b$ for $b$-bit expression while typical methods often choose $2^b$ values (\cite{Li2020}, \cite{Miyashita2016}, \cite{Zhou2017}).
Comparing to APoT, our method shares the idea of expanding the number of shifts.
However, while APoT only considers the addition of values, we also consider subtraction. 
Note that as well as the case of $n = 1$, the number of selected values is less than $2^b$.

Our approach considers introducing a bit-level sparsity in an efficient way; the number of selective values is expanded by considering subtraction.
Hence, the gap of the number of values between the uniform and proposed method becomes smaller as $b$ decreases.
This can be depicted quantitatively.
Equation \ref{eq:number_of_additive_values} presents the number of expressible values when only addition is considered for our proposed bit expression for a $b$-bit expression.
\begin{equation}
   N_{additive}(b, n) = {}_b C_n + {}_b C_{n-1} + ... + {}_b C_{1} + {}_b C_{0} 
   \label{eq:number_of_additive_values}
\end{equation}
When Equation \ref{eq:number_of_additive_values} is expanded to count the values that can be expressed by subtraction, the number of expressible values is expressed by Equation \ref{eq:number_of_nhot_values}, where $m$ denotes the number of continuous ``1''s. 
For simplicity, the case of $n=2$ is discussed here although a similar discussion can be made for other $n$.
\begin{align}
   &N_{n-hot}(b, n=2) = N_{additive}(b, n=2) + \sum_{m=3}^{b} b-m+1 \nonumber \\
   &\mbox{where $b\geq 3$}
   \label{eq:number_of_nhot_values}
\end{align}
From Equation \ref{eq:number_of_nhot_values}, we notice that when $b$ gets smaller, values that are expressed by subtraction decrease, and the number of expressible values approaches to $2^b$.
Hence, depending on the combination of $(b, n)$, $2^b$ values can be expressed by our proposed $n$-hot expression.
For instance, when $n = 2$ and $b = 3$, all values can be replaced by the $n$-hot quantization. 

\subsection{Overall training process}

When our proposed quantization scheme is applied, an accuracy drop may be induced due to the bit-level sparsity.
We observed that the accuracy drop occurs especially for object detection tasks that should be recovered.
To cope with this problem, a {\it two-stage fine-tuning algorithm} is used to recover the accuracy drop.
\vspace{2ex}
\begin{algorithm}[t]
   \caption{Two-stage fine-tuning algorithm}
   \label{alg:proposed_method}
   \begin{algorithmic}[1]
      \renewcommand{\algorithmicrequire}{\textbf{Input:}}
      \renewcommand{\algorithmicensure}{\textbf{Output:}}
      \REQUIRE{training data ${({\bf x_i}, y_i)}^L_{i=1}$, the number of minibatch $T$, 
      pre-trained weight ${\bf W_{full-precision}}$,
      scale $\alpha$, bit-width $b$, number of shift operations $n$, 
      warmup epochs $epoch_{warmup}$, total epochs $epoch_{total}$}
      \ENSURE{quantized weights ${\bf W_{n-hot}}$}


      [Activation quantization stage]
      \FOR{epoch = 1, 2, ..., $epoch_{warmup}$}
         \FOR{iter = 1, 2, ..., $T$}
         \STATE{Uniformly quantize activations}
         \STATE{Update $\bf W_{full-precision}$ by STE}
         \ENDFOR
      \ENDFOR

      [Weight quantization stage]
      \FOR{epoch = $epoch_{warmup}$+1, ..., $epoch_{total}$}
         \FOR{iter = 1, 2, ..., $T$}
         \STATE{Uniformly quantize activations}
         \STATE{Quantize weights by the proposed method\\
         ${\bf W_{n-hot}} \leftarrow \Pi_{Q_{n-hot}(\alpha, b, n)}({\bf W_{uniform}})$}
         \STATE{Update $\bf W_{n-hot}$ by STE}
         \ENDFOR
      \ENDFOR
   \end{algorithmic}
\end{algorithm}

{\bf Two-stage fine-tuning algorithm}: During training, comparing to full-precision networks, quantized networks are more likely to fall into local minima because of gradient approximation such as straight-through estimator (STE) (\cite{Courbariaux2016} and \cite{Hubara2016}).
Zhuang et al. \cite{Zhuang2018} indicated that appropriate initialization can prevent this problem and thus improve accuracy.
For this purpose, they proposed a two-stage optimization algorithm (e.g., firstly quantizing only weight and then secondly quantizing both weights and activations) and progressive quantization technique that gradually trains networks from higher to lower precision (e.g., 32-bit $\rightarrow$ 16-bit $\rightarrow$ 8bit $\rightarrow$ ...).
Similarly, we observed that appropriate initialization improves accuracy after fine-tuning for the training of object detection models. 
We used the two-stage fine-tuning algorithm similar to \cite{Zhuang2018} but in different optimization order. 
First, only activations are quantized and fine-tuned. 
Then, both activations and weights are quantized and fine-tuned by setting the weights trained at the first stage as initialization.

Algorithm \ref{alg:proposed_method} summarizes the overall training process.
In the experiment, $n$-hot quantization is applied for only weights and the quantization scheme for activations (lines 4 and 10) is the one given in Equation \ref{eq:minmax_quantization} (uniform quantization), where $m$ and $M$ denote the lower and upper limit of the quantization range, respectively and $x$ denotes the input data.
\begin{align}
   &\Pi_{Q_{uniform}(\alpha, b)}(x) \nonumber \\
   &= round \left(\frac{{min(max(x, m), M) - m}}{\alpha}
   \right) \times \alpha + m
   \label{eq:minmax_quantization} \\
   &\alpha = \frac{M - m }{2^b} \label{eq:dynamic_range}
\end{align}
The same scale is used for the $n$-hot quantization.
During the update steps (lines 5 and 12 in Algorithm \ref{alg:proposed_method}), STE is used to compute the gradients.

\section{Experiments}

\subsection{Experimental settings}

The experiments are performed on object detection and classification tasks by using the COCO \cite{COCO} and CIFAR100 \cite{Krizhevsky2008} datasets, respectively.
Our proposed method is compared with previously proposed methods, PoT, APoT, DoReFa-net \cite{Zhou2016}, and SBM \cite{Banner2018}.
For DoReFa-net and SBM, the experimental results are directly referenced from the study of Fu et al. \cite{Fu2021}.
For our proposed method, the experiment is conducted for the case of $n = 2$ because it has shown the best accuracy and computation cost trade-off.
For PoT and APoT, while their proposed quantization scheme is implemented, the same scale and backpropagation strategy as our proposed method are used.
They are denoted as PoT* and APoT*.
``Uniform'' means the quantization method given by Equation \ref{eq:minmax_quantization}.

The full-precision models of CenterNet with MobileNet-v2 and ResNet-18 backbone are trained by the same setting as the original paper but the different batch sizes; the batch sizes are 32 and 60 for MobileNet-v2 and ResNet-18 backbone, respectively.
MovileNet-v1 \cite{Howard2017} and -v2 \cite{Sandler2018} are fine-tuned for 800 epochs with the stochastic gradient descent (SGD) optimizer.
The learning rate at a certain epoch ($l_{epoch}$) is reduced by a cosine curve (Equation \ref{eq:cosine_scheduler}), where $l_{-1}$ denotes the initial learning rate and $\lambda$ is the period of the cosine curve.
\begin{equation}
   \label{eq:cosine_scheduler}
   l_{epoch} = l_{-1} \times \left(1 + cos\left(\frac{epoch}{\lambda}\pi \right)\right) 
\end{equation}
$l_{-1}$ is set as $1 \times 10^{-5}$ and 0.01 for object detection and classification, respectively except for non-pruned MobileNet-v1;
 $l_{-1}$ is 0.005 for non-pruned MobileNet-v1.
The weight decay is set as 0.0001.
Our proposed quantization scheme is also evaluated for pruned models.
The filters (e.g., channels) of the pre-trained networks are pruned by using the method of \cite{Molchanov2019}.
100 filters are pruned every 100 iterations unless the accuracy drop from the point of previous pruning is less than 0.1.
During the fine-tuning of the quantized models, the learning rate is scheduled by the cosine curve (Equation \ref{eq:cosine_scheduler}).
The initial learning rate is 0.001 and the weight decay is 0.0001.
For the object detection task, the activation and weight quantization stages in Algorithm \ref{alg:proposed_method} are both 80 epochs.
For the classification task, they are 20 and 60 epochs, respectively. 
The batch sizes are 8 and 32 for the object detection and classification tasks, respectively.

For activations, all activations except the last layer are quantized.
For weights, all convolution, depthwise, and deconvolution layers are quantized.
For CenterNet, not only the backbone network, but the detector layers are also quantized.
The batch normalization layers are folded into convolution layers.

The experiments are conducted on eight and four Tesla v100 GPUs running on an Ubuntu 18.04 operating system for the object detection and classification tasks, respectively.
Neural Network Libraries \cite{nnabla} is used for the implementation.


\begin{table*}[h]
\begin{center}
\begin{tabular}{|c|c|c|c|c|c|c|c|}
\hline
\multicolumn{5}{|c|}{CenterNet / backbone: MobileNet-v2} & \multicolumn{3}{|c|}{CenterNet / backbone: ResNet-18}\\
\hline
\begin{tabular}{c}Bit-width \\ (A / W)\end{tabular} & Method 
& \begin{tabular}{c}Accuracy \\ (mAP) \end{tabular} 
& \begin{tabular}{c}Model size \\ (MB)\end{tabular} 
& \begin{tabular}{c}bitOPs \\ (G)\end{tabular} 
& \begin{tabular}{c}Accuracy \\ (mAP) \end{tabular} 
& \begin{tabular}{c}Model size \\ (MB)\end{tabular} 
& \begin{tabular}{c}bitOPs \\ (G)\end{tabular} \\
\hline\hline
 32 / 32 & full precision & 0.232 & 45.5 & 26.8 & 0.251 & 65.4 & 22.1 \\
\hline
 8 / 8 & Uniform  & 0.225 & 12.2 & 2.79 & 0.244 & 18.2 & 1.38 \\
       & PoT*     & 0.177 & 12.2 & 0.348  & 0.228 & 18.2 & 0.178 \\
       & APoT*    & 0.226 & 12.2 & 0.697  & 0.246 & 18.2 & 0.346 \\
       & Proposed*& 0.222 & 10.8 & 0.697  & 0.243 & 16.4 & 0.346 \\
       & {\bf Proposed} & {\bf 0.224} & {\bf 10.8} & {\bf 0.697}  & {\bf 0.245} & {\bf 16.4} & {\bf 0.346} \\
\hline
 8 / 6 & Uniform  & 0.223 & 10.8 & 2.09   & 0.242 & 16.4 & 1.09 \\
       & PoT*     & 0.175 & 10.8 & 0.348  & 0.228 & 16.4 & 0.173 \\
       & APoT*    & 0.219 & 10.8 & 0.697  & 0.243 & 16.4 & 0.346 \\
       & Proposed*& 0.219 & 10.1 & 0.697  & 0.242 & 15.5 & 0.346 \\
       & {\bf Proposed} & {\bf 0.220} & {\bf 10.1} & {\bf 0.697}  & {\bf 0.244} & {\bf 15.5} & {\bf 0.346} \\
\hline
 6 / 6 & Uniform  & 0.193 & 9.44 & 1.57   & 0.235 & 14.3 & 0.779 \\
       & PoT*     & 0.151 & 9.44 & 0.261  & 0.222 & 14.3 & 0.130 \\
       & APoT*    & 0.193 & 9.44 & 0.522  & 0.236 & 14.3 & 0.260 \\
       & Proposed*& 0.191 & 8.73 & 0.522  & 0.231 & 13.4 & 0.260 \\
       & {\bf Proposed} & {\bf 0.209} & {\bf 8.73} & {\bf 0.522}  & {\bf 0.235} & {\bf 13.4} & {\bf 0.260} \\
\hline
\end{tabular}
\end{center}
\caption{Results for COCO on different bit-width, method, accuracy, model size, and bitOPs on CenterNet (backbone MobileNet-v2 and ResNet-18) models.}
\label{tab:COCO_results}
\end{table*}

\subsection{Experimental results for COCO}

The performance of different quantization methods is compared for their accuracy, model size, and the number of binary operations (bitOPs) under different bit-widths.
The sign and absolute value bit are 1-bit and $b$-bit, respectively. 
To evaluate the number of bitOPs, the number of multiplication operations between activations and weights is counted.
For instance, while the multiplication between $b_a$-bit activation and  $b_w$-bit weight requires $b_a \times b_w$ bitOPs, the multiplication between $b_a$-bit activation and a weight with $n$ ``1'' bit values requires $b_a \times n$ bitOPs.
The theoretical model size is calculated by considering the compression rate of each layer. 
The accuracy metric for the COCO dataset is the mean average precision (mAP) for intersection over union (IoU) between 0.5 and 0.95.

Table \ref{tab:COCO_results} summarizes the experimental results of the COCO dataset. 
The proposed method is evaluated with and without (denoted as ``proposed'' and ``proposed*'' in Table \ref{tab:COCO_results}) the two-stage fine-tuning algorithm. 
Compared to the uniform method, the proposed* method suppresses the accuracy drop within 0.3\% while reducing the bitOPs by about 75\% and the model size by 11.5\% (8 / 8-bit, MobileNet-v2 backbone). 
Similar accuracy is achieved for the 8 / 8-bit proposed method and 8 / 6-bit uniform method. 
This infers that while our proposed method reduces the number of quantization levels of an 8-bit model as the same as a 6-bit model in a non-uniform way, the recognition performance is kept at the level of the uniform model. 
The proposed two-stage fine-tuning algorithm effectively improves the accuracy. 
For instance, the accuracy is improved by 1.8\% and 0.4\% for MobileNet-v2 and ResNet-18 backbone models, respectively (6 / 6-bit). 

Compared to the uniform method, the accuracy drop for PoT* is severe (4.8\% for 8 / 8-bit, MobileNet-v2). 
Because CenterNet model is sensitive to quantization, the accuracy degradation for PoT may be severe because the quantization levels near the maximum value are sparse.
APoT * achieves similar accuracy to the uniform with the reduced bitOPs. 
However, our proposed method with the two-stage fine-tuning achieves a similar or even better accuracy than APoT and further reduces the model size by 11.5\% (8 / 8-bit, MobileNet-v2 backbone).

Note that the network is fully quantized; not only the backbone network but the detector network including computationally expensive deconvolution layers are also quantized. The experimental results indicate that our proposed method can effectively compress sensitive object detection models while keeping the accuracy.

\subsection{Experimental results for CIFAR100}

\begin{table}
\begin{center}
\begin{threeparttable}
\begin{tabular}{|c|c|c|c|c|}
\hline
\multirow{2}{*}{Method} & \multicolumn{2}{|c|}{MobileNet-v2 \tnote{1}} &\multicolumn{2}{|c|}{MobileNet-v1\tnote{2}} \\
\cline{2-5}
 & 8 / 8-bit & 6 / 6-bit & 8 / 8-bit & 6 / 6-bit\\
\hline\hline
DoReFa  & 70.2 & 68.6 & - & - \\
SBM     & 75.3 & 72.3 & - & - \\
Uniform & 73.9 & 73.1 & 73.1 & 71.7 \\
APoT*   & 73.9 & 73.1 & 72.7 & 71.4 \\
PoT*    & 71.8 & 71.6 & 71.1 & 70.1 \\
{\bf Proposed*} & {\bf 73.5} & {\bf 72.7} & {\bf 72.7} & {\bf 71.3}\\
\hline
& \multicolumn{2}{|c|}{\begin{tabular}{c}42.0\% channel \\pruned \end{tabular}}
& \multicolumn{2}{|c|}{\begin{tabular}{c}93.6\% channel \\pruned \end{tabular}}\\
\hline
Uniform & 72.2 & 71.8 & 61.1 & 61.5 \\
APoT*   & 72.3 & 71.6 & 61.0 & 61.3 \\
PoT*    & 71.0 & 70.8 & 59.4 & 58.8 \\
{\bf Proposed*} & {\bf 72.3} & {\bf 71.6} & {\bf 61.5} & {\bf 61.3}\\
\hline
\end{tabular}
\begin{tablenotes}
   \item[1] Full precision accuracy: 72.5\%
   \item[2] Full precision accuracy: 72.2\%
\end{tablenotes}
\end{threeparttable}
\end{center}
\caption{Accuracy (\%) for CIFAR100 on different methods on MobileNet-v2 and v1.}
\label{tab:CIFAR100_results}
\vspace{2ex}
\end{table}

Table \ref{tab:CIFAR100_results} summarizes the accuracy of the CIFAR100 dataset.
Similar characteristics to the case of object detection are observed. 
Our proposed method suppresses the accuracy loss by 0.4\% (MobileNet-v2, 6 / 6-bit at most) while PoT* suffers from accuracy loss because of its sparse quantization levels. 
However, SBM outperforms other methods in the case of 8 / 8-bit. 
This infers that introducing Range BN and differentiable backward function may improve accuracy for 8-bit networks.

For pruned models, the accuracy loss for the proposed method compared to other methods is small as well as non-pruned situations (0.4\% for MobileNet-v1, 8 / 8-bit at most). 
For the pruned models of MobileNet-v1, the results show the different characteristics to others; the accuracies are similar between 8 and 6-bit precisions.
This infers that the accuracy loss may be induced mostly by aggressive pruning.
The accuracy loss by $n$-hot quantization is not obvious as well.
The results indicate that our proposed $n$-hot quantization scheme is robust to pruning.
In the combination with pruning, further model compression can be achieved without a large accuracy loss that is triggered by quantization.

\subsection{Ablation studies}

\begin{table}
\begin{center}
\begin{threeparttable}[h]
\begin{tabular}{|c|c|c|c|c|}
\hline
\multirow{2}{*}{Method}
&\multicolumn{2}{|c|}{\begin{tabular}{c}CenterNet \\ MobileNet-v2, \\mAP \end{tabular}} 
&\multicolumn{2}{|c|}{\begin{tabular}{c}CenterNet \\ ResNet-18, \\mAP \end{tabular}} \\
\cline{2-5}
& \begin{tabular}{c}8 / 8\\-bit \end{tabular} 
& \begin{tabular}{c}6 / 6\\-bit \end{tabular} 
& \begin{tabular}{c}8 / 8\\-bit \end{tabular} 
& \begin{tabular}{c}6 / 6\\-bit \end{tabular} \\
\hline\hline
\begin{tabular}{c}PoT*\\($n=1$)\end{tabular}& 0.177 & 0.151 & 0.228 & 0.222 \\
\hline
\begin{tabular}{c}One-hot\tnote{1}\\($n=1$)\end{tabular}& 0.182 & 0.144 & 0.229 & 0.211 \\
\hline
\begin{tabular}{c}Addition\\($n=2$)\end{tabular}& 0.221 & 0.187 & 0.242 & 0.229 \\
\hline
\begin{tabular}{c}{\bf Addition \&} \\{\bf subtraction}\\($n=2$) \end{tabular}
&\bf{0.222} & \bf{0.191} & \bf{0.243} & \bf{0.231} \\
\hline
\end{tabular}
\begin{tablenotes}
   \item[1] Special case of PoT* that the number of bit values is $b$
\end{tablenotes}
\end{threeparttable}
\end{center}
\caption{Evaluation (mAP) of the proposed method for different $n$ and subtraction on CenterNet models on COCO.}
\label{tab:ablation_studies_results}
\vspace{2ex}
\end{table}

Our proposed $n$-hot bit representation takes the parameter $n$.
Moreover, it proposes to expand representative values by considering subtraction of PoT terms.
In this section, the analysis of the parameter $n$ and the effect of considering subtraction is discussed.
Table \ref{tab:ablation_studies_results} summarizes the accuracy (mAP) of the proposed method for $n=1, 2$ and different bit expressions for CenterNet models on COCO dataset.
``PoT*'' and ``one-hot'' denote the case of $n=1$, but the numbers of weight values are $2^b$ and $b$, respectively.
``Addition'' and ``Addition \& subtraction'' are the bit representation of the proposed method but denote the cases of only considering an addition or both addition and subtraction, respectively.
For this experiment, the two-stage fine-tuning algorithm is not applied.

Comparing to the cases of $n=2$, the accuracy decreases for the case of $n=1$ (PoT* and one-hot).
As discussed in the previous section, $n=1$ bit representation suffers from expression ability and leads to accuracy degradation.
However, the accuracy is similar for PoT* and one-hot.
This implies that although PoT* has $2^b$ values, the expression ability is poor. 
Considering the accuracy and model size trade-off, if $n=1$ architecture is chosen, the one-hot representation may be the best option because the model compression rate is much larger.

Comparing ``Addition'' and ``Addition and subtraction'', the accuracy is better for ``Addition and subtraction'' for all bit-widths and models.
The accuracy is 0.4\% and 0.2\% better for MobileNet-v2 and ResNet-18 backbone models, respectively for 6 / 6-bit.
Therefore, considering the subtraction is important to improve the accuracy of the proposed $n$-hot quantization method. 

\section{Conclusion}

In this paper, we propose an efficient non-uniform quantization scheme that balances the accuracy and computation costs.
Our proposed $n$-hot quantization scheme introduces bit-level sparsity into the bit representation; values are expressed by either addition or subtraction of $n$ PoT values.
Also, we show the improvement in the performance of our proposed quantization scheme by using the two-stage fine-tuning algorithm. 
The evaluation is performed for classification and object detection tasks. 
The experimental results show that our proposed method suppresses the accuracy drop by 0.3\% at most while reducing the bit operations by about 75\% and model size by 11.5\% compared to the uniform method for an 8-bit CenterNet model on the COCO dataset.
Similarly, the proposed quantization scheme suppresses the accuracy drop by 0.4\% at most for the CIFAR100 dataset.
Our proposed quantization scheme is also evaluated for pruned networks and showed robustness.
We hope that our approach facilitates the implementation of DNN to resource-constrained HW.

{\small
\bibliographystyle{ieee_fullname}
\bibliography{egpaper_for_review}
}

\end{document}